# Tensity Research Based on the Information of Eye Movement

Yi Wang[1] and Zhiliang Wang[2]

[1]School of Automation and Electrical Engineering, University of Science and Technology Beijing,
Beijing, China, 100083

[2]School of Computer and Communication Engineering, University of Science and Technology Beijing,
Beijing, China, 100083

**Abstract**

User's mental state is concerned gradually, during the interaction course of human robot. As the measurement and identification method of psychological state, tension, has certain practical significance role. At presents there is no suitable method of measuring the tension. Firstly, sum up some availability of eye movement index. And then parameters extraction on eye movement characteristics of normal illumination is studied, including the location of the face, eyes location, access to the pupil diameter, the eye pupil center characteristic parameters. And with the judgment of the tension in eye images, extract exact information of gaze direction. Finally, through the experiment to prove the proposed method is effective.

## 1. Introduction

By watching eyes can know thoughts. Eyes are not only the window of us knows the world and the routine of communication, it reacts within us the window of the world. In recent decades, (eye tracking) the development of the line of sight technology allows users to more accurate record of eye movement indicators, by which we explore the index's relationship with human cognitive psychology.

The main task of cognitive psychology is to research all kinds of psychological phenomenon and the cognitive process, and tries to reveal the humanity is how to obtain information from the outside world, and how the information processing, code, and stored in the brain. To be precise, it mainly includes the sense perception, attention, memory, thinking, imagination, language, and other psychological phenomenon and process [7].

In addition to the early traditional tension measurement indicators, such as: skin resistance, blood pressure, heart rate, etc. appeared in the 1970s, the United States according to the phonetic index to measure the MARK - comprehensive analysis of the polygraph. This is based primarily on the principle of human nervous muscle tremors developed at different frequency, within which 90% results is accurate [10].

In 1987, some researchers for the first time using event related potential (ERP) in the brain electrical measuring tension. Currently utilizing ERP P300, the event stimulus time is set around 300 ms after is sine waveform appeared. The indexes including the analysis of the amplitude, the incubation period and scalp distribution [7].Domestic study of P300 detector is relatively small, mainly used in medicine.

Functional magnetic resonance imaging (fMRI) technique is a direct detection stressed human brain areas of the excited state of technology; the first is to study the Pennsylvania. But the research of this technology is currently in phase, needs further verification. Pupil diameter index, early as possible measures tension indicators into the researcher's research scope. In 1933 the company found strong emotions can make pupil dilation, but the result of the experiment is very vague and less [9].Heilveil [10] in 1976 to join in the test for the pupil diameter measurement, the pupil diameter amplification of most of the participants is considered very nervous. Dionysus Granholm, Hillix and Perrine [18] scenes on the subjects and language information test, record the tension when the subjects' pupil diameter. Experiments have established that liar pupil diameter change is big, but for the test of the information of the two different and no difference. The researchers found in the comparison also pupil diameter index and skin pressure index has the exact same effect [2].On the basis of these studies, we can see that the pupil diameter can be used as the index. Some special indeed utilized for measuring the tension.

## 2. Characteristic Parameter Extraction Method for the Eyes

### 2.1 REM Measurement Method

Eye movement tracks measurement, also known as the line of sight, its purpose is to record the eye gaze direction on the screen. First eye movement record method is through direct observation; development so far, there have been a lot of kinds of records of method, can be divided into two kinds of invasive and noninvasive. This article selected is invasive based on digital Video, Video Oculographic,

VOG) of eye movement record method, this method can be divided into the following categories.
The Pupil and The corneal Reflection method (The Pupil Center Cornea Reflection Technique, PCCR): use infrared light source is in The eye of Sipkhin spot, as The benchmark for The eye movement, using bright dark Pupil difference method to obtain The Pupil image. Using the pupil center and vector of Sipkhin to estimate the line of sight [6].
The corneal reflection matrix Method (Cross - Ratios Method) [8]: using multiple infrared light sources produced in the cornea corneal reflection matrix of data structure, adopting this position to estimate the pupil center corneal reflection matrix and the line of sight.
Method of ellipse normal direction (" One Circle "Algorithm): using a high-resolution camera to gather the eye image, obtain the boundary of the iris of the normal direction to estimate the line of sight [4]. This paper aims to adopt a more convenient way for eye line of sight, also hope to explore how to under the condition of light source often get eye movement indicators. So on pupil - corneal reflection method (PCCR) were studied after the decision under the condition of light source often use the pupil - corneal reflection principle, adopt the pupil - method to get the corner of my eye gaze. PCCR using infrared light source on the corneal reflection of Sipkhin spot as a reference of eye movement. But the constant light gathering Sipkhin spot, this article choose can thus as reference of eye movement, fixed constraints is the person's head, and face the screen.

## 2.2 The Method of Extracting Characteristic Parameters

In this paper, the feature vector refers to the characterization of the line of sight direction parameter. Traditional takes two steps to implement the method of the cornea and pupil, the feature parameter extraction and the line of sight direction map. In order to make sure that the subjects gaze point on the screen, then you must know the subjects' gaze direction.
Sweden famous ophthalmology expert Allvar Gullstrand put forward a mathematical description of eyeball structure model, and gives the light from the air, through the cornea, lens and so on several different refractive index medium, and in the retina imaging principle, as shown in table 1, lists several major membrane layer structure and some basic indicators.[ 1]
Gullstrand research achievements of scholars in the field of general acceptance, and the model named Gullstrand eye model. At present many researchers in the field of vision to track this eye model on the basis of the specific representation of the line of sight direction. Many researchers [1] [2] Gullstrand eyeball is given a detailed description of the model. As shown in figure 1 is given in literature [41] that the model and the line of sight direction. Representation method is shown in formula (1), according to a few basic physiological indicators in the main membrane layer structure of eye.

Table 1: Basic physiological indicators of several major film structures in the eye

|  | Location(mm) | Radius(mm) | Refractive index via medium |
|---|---|---|---|
| cornea | 0 | 7.7 | 1.376 |
|  | 0.5 | 6.8 | 1.336 |
| lens | 3.2 | 5.33 | 1.385 |
|  | 3.8 | 2.65 | 1.406 |
|  | 6.6 | -2.65 | 1.385 |
|  | 7.2 | -5.33 | 1.336 |
| retina | 24.0 | -11.5 |  |

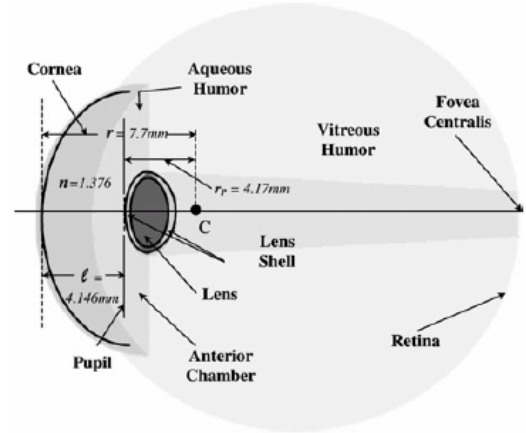

Fig. 1 Gullstrand eye model.

The direction of the line of sight can be expressed as :

$$\bar{E} = (p - c)/r_p \qquad (1)$$

Where, $\bar{E}$ is the line of sight direction vector; P is the pupil center coordinates (3D);C is eye center coordinates (3D); $r_p = \|p - c\|$.

According to the direction of the sight line, expression is introduced to the center of the pupil, by which to describe the main characteristic of the information. Followed by the type of description in the eye center, the relative static reference is settled. In PCCR approach does not directly measure the center of the eye but determined using infrared light on the Sipkhin flare point of cornea to as a reference, that on the premise of corneal approximation for spherical, Sipkhin of the relative position is the same. As shown in the figure below:

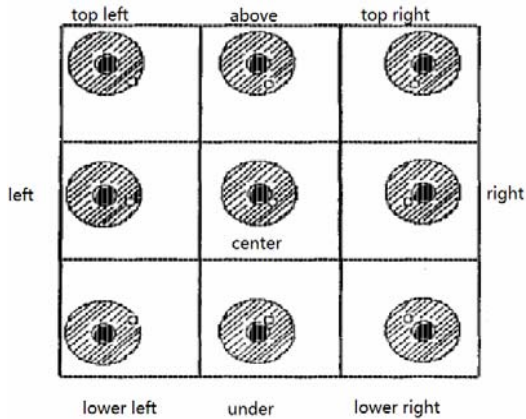

Fig. 2 Eye movement to a different pupil position and the relative position of the Sipkhin spot.

As described in the picture, when the eye moves to nine different position, the relative position of pupil and Sipkhin spot, of which the shadow is part of the iris, the black part is a pupil, white small circle is &poor's Sipkhin be seen in the figure, although their relative position has changed, but Sipkhin of absolute position has not changed, so can be Sipkhin spot as a pupil change reference quantity, without measuring the center of the eye, because this feature is not obvious. In a conventional light source system, the original Sipkhin spot replacing corner points, because on the premise of the head rest, a corner of absolute position is the same. Eyes characteristic parameters of extraction process, as shown in the figure below:

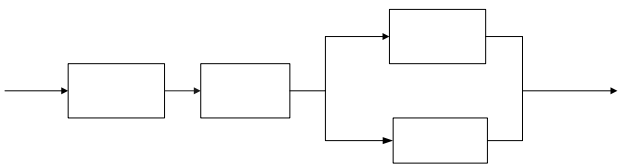

Fig. 3 Process of Eye parameters detection.

## 3. The Human Eye Detection

Before to extract the characteristic parameters of the eye, you first need to on the video image detection to the images of the human eye, but before the eyes of image need to detect human face image. So we can narrow the scope of the image processing, and improve the detection accuracy and efficiency.

Face detection belongs to the target detection, object detection), as part of the mainly involves two aspects:

(1) first to probability and statistics, to detect the target object to know some of the characteristics of the object to be detected, established a model of the target detection.
(2) model applied to match the input image, if there are matching output matching area, otherwise don't do anything.

This article uses Opencv classifier. of Harr features. It is first designed by Paul Viola and Michael Jones, known as the Viola-Jones detector. It then by Rainer Lienhart and Jochen Maydt expand with diagonal characteristics.[43]. It uses the weak classifier Adaboost algorithm of node selection in cascade, in which each node is composed of multiple trees, for the most part just a decision tree, in the form of a layer of decision tree to allow the following decisions: v determine characteristics of f value is greater than a certain threshold t; 'Yes' said may be face, 'no' said not face:

$$f_i = \begin{cases} +1, v_i \geq t_i \\ -1, v_i < t_i \end{cases}.$$

In training, Viola - Jones classifier in each weak classifier using class Harr number - like features can be set. But in most cases, the application of one feature, at most no more than three features. Then, boosting algorithm, iteratively build a composed of those weak classifier weighted and strong classifier. The classification of the Viola-Jones, use the following functions:

$$F = sign(\sum_i w_i f_i).$$

If the weighted and less than zero, the function returns 1;Equal to zero, it returns 0;Greater than zero, it returns 1.1 does not match, 0 means not matching, 1 means to match. Each node of the correct recognition rate is very high, but the correct rejection rate is very low (that is, a lot of face be detected).At any level of computing, once obtained the goal, is shown as "not in the category" conclusion, computation terminates immediately. This point is when the target frequency is low (for example, a lot of images in only a small pair of small face), screening of cascade classifier can significantly reduce the amount of calculation. Because most of the tested area can be early prepared, or by the filter, quickly identified here with or without facial image.

Opencv Harr feature is used in class function, small class precisely Harr potter. Detection using the cascade table of haar features, the cascade is contained in the boost of classifier. First, people using harr features of samples to the training of the classifier, to get a boost of cascade classifier. Training includes two aspects:
1. Are samples, namely the target samples to be detected.
2. The cases of samples, pictures of any other samples.
To unite these images to the same size, a process called normalization, and statistics. Once classifier to establish complete, it can be used to detect the input images of interested area, in general, images will input is greater than the sample, so, need to move the search window, in order to retrieve the goals out of different sizes, classifier can change their size in proportion, so may be to many times of scanning the input image.

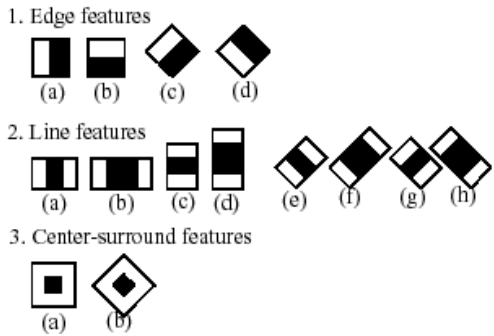

When scanning images to be detected in the boundary characteristics, (a), for example, as mentioned earlier as a result, the computer in the picture is composed of a digital matrix, first calculate the gray value of the x window, and then calculate the rectangular box black grey value of y, then calculating the value of (x-2y), comparing the numerical and eventually x, in a certain range, if the ratio is said to be detected image of the current scan area (a) conform to the boundary characteristics, and then continue to scan.

When according to the principle of the above selected face feature, then will face as input, detection of the human eye. The human eye detection principle is similar to the principle of face detection, also use Harr trainers.

## 4. The Pupil Parameters Testing

Considering the pupil hidden many inner psychological characteristics of information, its relationship with the polygraph cognitive science have explain clearly in the previous chapter. This paper involves the pupil parameters mainly include: the center of the pupil detection and the radius of the detection of the pupils, which center can be used as tension judgment direct basis, radius of the pupil of eye movement parameters detection, can be used for indirect tension judgment basis.

The pupil center detection method in this paper draw lessons from the part of the principle and steps of iris recognition. Iris recognition of the need to detect the edge and outside edge, the edge is within the boundary of the pupil.

Pupil boundary approximation is circular, but due to differences in image acquisition device, structure of different human eyes, the eyes of the activities and the interference of eyelid and eyelashes, collected the pupil image often is not round. So the pupil orientation can be thought of as a round matching problem.

In the operator of the iris localization, Daugman J of the scale of the blur of the effect of differential integral operator is optimal. But this algorithm needs several times for each pixel of the image of the differential points and convolution operation, the computation is very large, for real-time systems, is inappropriate. Richard P.W ides after Dougman two-step method was proposed, its history of binary image first, then apply Hough detection round operator, positioning inside and outside the boundary of the pupil, time is greatly reduced.But the Hough transform retained a large amount of redundant information, so the positioning time still cannot meet the requirements of real-time systems.

Based on the automatic threshold segmentation pupil detection method in this paper, Due to the unique characteristics of the pupil, iris, and sclera of the grey value is increased in turn. For Asian people and the pupil and iris gray level difference is bigger, so according to the accuracy of grey value to distinguish the pupil is relatively high. The pupil center location method based on threshold segmentation flow chart is as follows:

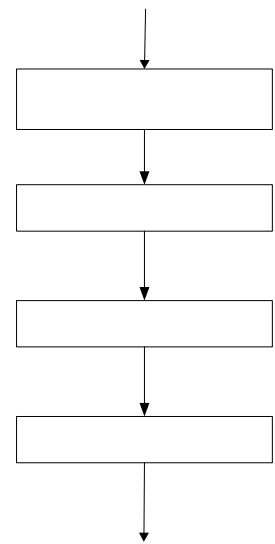

Fig. 4 Searching method of pupil center location based on threshold segmentation.

(1) Automatic selection threshold image segmentation
Select the appropriate threshold can accurately segment the pupil area, but the area will still contain the eyelash or portions of the eyelids. The selection of threshold is core eye image gray histogram. Gray histogram effectively reflects the image in a gray or the number of occurrences of frequency.

Of image gray histogram, we first use formula (2), using integer Gaussian filter to smooth the gray-level histogram.

$$h(t) = \frac{\exp(-t^2/2\delta^2)}{\sqrt{2\pi}\delta}, \delta = \frac{\sqrt{\ln 2}}{2\pi BT} \quad (2)$$

The pupil area location

After the automatic threshold segmentation image, get the binary image in addition to clear the pupil area will be mixed with other area of the noise, such as palpebral eyelash or due to the light and the dark area. Adopt the method of morphological filtering can eliminate the noise in the image and information as well as some interference, and can be carried out on the boundary point filling and repair, to fill the "empty".

Due to the pupil image closer to the circle, so this paper selected circular template, radius is 3 mm , radius is eroding the pupil area, is too small will cause corrosion expansion too many times. Template se shape is as follows:

$$se = \begin{bmatrix} 0 & 0 & 0 & 1 & 0 & 0 & 0 \\ 0 & 0 & 1 & 1 & 1 & 0 & 0 \\ 0 & 1 & 1 & 1 & 1 & 1 & 0 \\ 1 & 1 & 1 & 1 & 1 & 1 & 1 \\ 0 & 1 & 1 & 1 & 1 & 1 & 0 \\ 0 & 0 & 1 & 1 & 1 & 0 & 0 \\ 0 & 0 & 0 & 1 & 0 & 0 & 0 \end{bmatrix}.$$

In this paper, the morphological filtering method as shown in the following formula (3):

$$f' = erode^{N2-N1}\{dilate^{N2}(erode^{N1}(\sim f))\} \quad (3)$$

Where, $f$ is the original binary image, $f'$ is image after the morphological filtering. The Formula (3) can be described as follows, the first to take the binary image. For the reason of the morphological filtering is effective in white spots. After the next to take the image of N1 corrosion formula (4),

$$f \odot b(s,t) = \min\{f(s+x,t+y)+b(x,y)|(s+x),(t+y) \in D_f;(x,y) \in D_h\} \quad (4)$$

The function of the corrosion is to eliminate the noise in the image. In the noise points including some bright spots, difference in palpebral eyelash and some darker areas. Through such processing, maximum limit retained the pupil area. Then for the expansion of the formula (4), N2 times operations.

$$f \oplus b(s,t) = \max\{f(s-x,t-y)+b(x,y)|(s-x),(t-y) \in D_f;(x,y) \in D_h\} \quad (4)$$

After positioning in the pupil area, namely after the morphological filtering, basic the circular area of the pupil is determined. This article adopts the method of pixel statistics to determine the location of the center of the circle. Circle can be considered through the graphics of the longest two lines of intersection point that is the center of the pupil. $(x_0, y_0)$ is the biggest rows and columns of the image pixel point, using estimates for the number of pixels radius, as shown in formula (6) :

$$\begin{cases} x_0 = row_{max} \\ y_0 = col_{max} \\ r = \sqrt{\dfrac{num_{pix}}{\pi}} \end{cases} \quad (6)$$

Based on support vector machine model, it can be implemented to the line of sight coordinate judgment, as shown in the figure below:

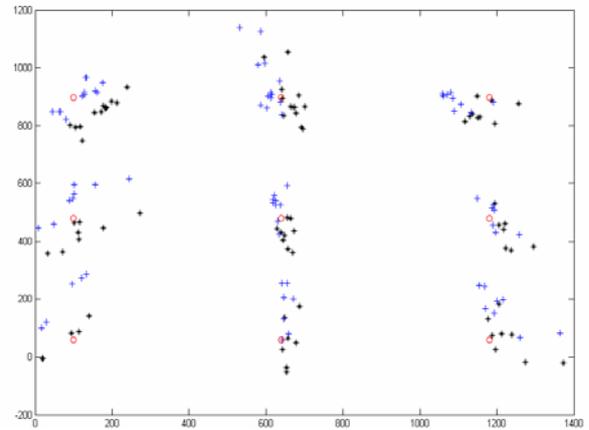

Nervous trigger physiological reaction. This hypothesis, when really accept psychological tests, will be aware of their own problems, it will produce nervous heart, will also increase the cognitive load.

Different attention to the problem, innocent people focus on the question of the criterion, laying focus on related issues. With people being tested on the criterion and the related problems of psychological response difference to judge or tested for issues related to the existence of abnormal psychological pressure.

## 5 Conclusion

This paper studied eye movement. First of all, it browsed through a large number of reading materials, then it summarized some availability of eye movement indicators. Subsequent to often subtle eye movement characteristic, parameters extraction is studied, including the positioning faces, locate the human eye, pupil diameter, eye pupil center characteristic parameters. And on this basis, the extraction of line of sight direction discriminates for tension.

**Acknowledgments**

This work is supported by National Natural Science Foundation of China (No. 61170117).

**References:**

[1] Wildes P.Iris, "Recognition：An Emerging Biometric Technology", Proceedings of the IEEE，1997，85(9)：1348-1363.

[2] Y. B. Tian, Z. H. Xu, H. J. Feng, et al. "Pupilcontrolled auto2focusing technique used in digital cameras", Optical Technique, 2003, 29(1): 53255.

[3] D. Beymer and M. Flickner, "Eye gaze tracking using an active stereo head," in Proc. CVPR, Vol. 2, 2003, pp. 451–458.

[4] S. W. Shih and J. Liu, "A novel approach to 3-D gaze tracking using stereo cameras," IEEE Trans. Syst. Man Cybern. B, Vol. 34, no. 1, 2004, pp.234–245.

[5] T. Ohno, N. Mukawa, and A. Yoshikawa, "Freegaze: A gaze tracking system for everyday gaze interaction," in Proc. Symp. ETRA 2002,2002, pp. 125–132.

[6] H. Morimoto, A. Amir, and M. Flickner, "Detecting eye position and gaze from a single camera and 2 light sources," in Proc. Int. Conf.Pattern Recognition, 2002, pp. 314–317.

[7] Y. Matsumoto, T. Ogasawara, and A. Zelinsky, "Behavior recognition based on head pose and gaze direction measurement," in Proc. 2000IEEE/RSJ Int. Conf. Intell. Robots Syst., 2000, pp. 2127‐2

[8] R. E. Lubow, and Ofer Fein, "Pupillary Size in Response to a Visual Guilty Knowledge Test: New Technique for the Detection of Deception", Journal of Experimental Psychology: Applied, Vol. 2, No. 2, 1996, pp.164-177.

[9] A. Vrij, Detecting lies and deceit: Pitfalls and opportunities (2nd ed.). Chichester, England: Wiley, 2008.

[10] D. Kahneman, and J. Beatty, "Pupil diameter and load on memory", Science, Vol.154, 1996, pp.1583‐1585.

**First Author** Yi Wang is a PhD student at School of Automation and Electrical Engineering, University of Science and Technology Beijing. His current research interests include Humanoid Robot and Artificial Psychology .

**Second Author** Wang Zhiliang is chief professor of University of Science and Technology Beijing, and member of its Academic Committee. His current research projects include: the Digital Technology and Research of Artificial Psychology, Face Recognition, Facial Expression Recognition, Gait Recognition, Commentary Based Service Robot, Individualized Web Information Service System, Digital Entertainment and Entertainment Education.